\pdfoutput=1
\documentclass[10pt,twocolumn,letterpaper]{article}

\usepackage{cvpr}
\usepackage{times}
\usepackage{epsfig}
\usepackage{graphicx}
\usepackage{amsmath}
\usepackage{amssymb}

\usepackage{enumerate}
\usepackage{graphicx}
\usepackage{array}
\usepackage[lofdepth,lotdepth]{subfig}
\newcommand\numberthis{\addtocounter{equation}{1}\tag{\theequation}}

\newcolumntype{L}[1]{>{\raggedright\let\newline\\\arraybackslash\hspace{0pt}}m{#1}}
\newcolumntype{C}[1]{>{\centering\let\newline\\\arraybackslash\hspace{0pt}}m{#1}}
\newcolumntype{R}[1]{>{\raggedleft\let\newline\\\arraybackslash\hspace{0pt}}m{#1}}

\graphicspath{ {fig/}{fig/edge_map/}{fig/sk_ed/}{fig/multi_cls/}{fig/all/} }
\usepackage[pagebackref=true,breaklinks=true,letterpaper=true,colorlinks,bookmarks=false]{hyperref}

\cvprfinalcopy 

\def\cvprPaperID{2335} 

\ifcvprfinal\fi
\begin{document}
\setlength{\textfloatsep}{10pt}
\setlength{\abovedisplayskip}{4pt}
\setlength{\belowdisplayskip}{4pt}

\title{SketchyGAN: Towards Diverse and Realistic Sketch to Image Synthesis}

\author{Wengling Chen\\
Georgia Institute of Technology\\
{\tt\small wchen342@gatech.edu}
\and
James Hays\\
Georgia Institute of Technology, Argo AI\\
{\tt\small hays@gatech.edu}
}

\twocolumn[{%
\renewcommand\twocolumn[1][]{#1}%
\maketitle
\vspace*{-3em}
\begin{center}
\centering
\includegraphics[width=1.0\textwidth]{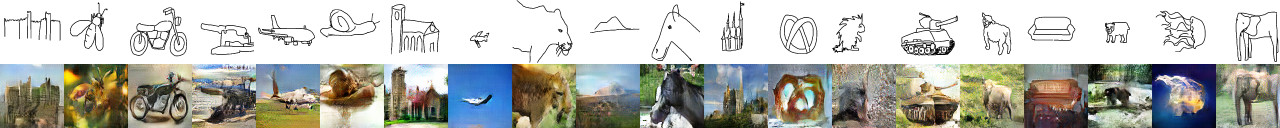}
\captionof{figure}{A sample of sketch-to-photo synthesis results from our 50 categories. Best viewed in color.}
\label{fig:final_output_abs}
\end{center}
}]

\begin{abstract}
   Synthesizing realistic images from human drawn sketches is a challenging problem in computer graphics and vision. Existing approaches either need exact edge maps, or rely on retrieval of existing photographs. In this work, we propose a novel Generative Adversarial Network (GAN) approach that synthesizes plausible images from 50 categories including motorcycles, horses and couches. We demonstrate a data augmentation technique for sketches which is fully automatic, and we show that the augmented data is helpful to our task. We introduce a new network building block suitable for both the generator and discriminator which improves the information flow by injecting the input image at multiple scales. Compared to state-of-the-art image translation methods, our approach generates more realistic images and achieves significantly higher Inception Scores.\footnote{Code can be found at \href{https://github.com/wchen342/SketchyGAN}{https://github.com/wchen342/SketchyGAN}}
\end{abstract}


\section{Introduction}
How can we visualize a scene or object quickly? One of the easiest ways is to draw a sketch. Compared to photography, drawing a sketch does not require any capture devices and is not limited to faithfully sampling reality. However, sketches are often simple and imperfect, so it is challenging to synthesize realistic images from novice sketches. Sketch-based image synthesis enables non-artists to create realistic images without significant artistic skill or domain expertise in image synthesis. It is generally hard because sketches are sparse, and novice human artists cannot draw sketches that precisely reflect object boundaries. A real-looking image synthesized from a sketch should respect the intent of the artist \emph{as much as possible}, but might need to deviate from the coarse strokes in order to stay on the natural image manifold. In the past 30 years, the most popular sketch-based image synthesis techniques are driven by image retrieval methods such as Photosketcher \cite{Eitz11} and Sketch2photo \cite{chen2009sketch2photo}. Such approaches often require carefully designed feature representations which are invariant between sketches and photos. They also involve complicated post-processing procedures like graph cut compositing and gradient domain blending in order to make the synthesized images realistic.\\\indent
The recent emergence of deep convolutional neural networks \cite{lecun2015DeepLearning,Krizhevsky14AlexNet,he2016ResNet} has provided enticing methods for image synthesis, among which Generative Adversarial Networks (GANs) \cite{Goodfellow14} have shown great potential. A GAN frames its training as a zero-sum game between the generator and the discriminator. The goal of the discriminator is to decide whether a given image is fake or real, while the generator tries to generate realistic images so the discriminator will misclassify them as real. Sketch-based image synthesis can be formulated as an image translation problem conditioned on an input sketch. There exist several methods that use GANs to translate images from one domain to another \cite{pix2pix2016,CycleGAN2017}. However, none of them is specifically designed for image synthesis from sketches.\\\indent
In this paper, we propose SketchyGAN, a GAN-based, end-to-end trainable sketch to image synthesis approach that can generate objects from 50 classes. The input is a sketch illustrating an object and the output is a realistic image containing that object in a similar pose. This is challenging because: (i) paired photos and sketches are difficult to acquire so there is no massive database to learn from. (ii) There is no established neural network method for sketch to image synthesis for diverse categories. Previous works train models for single or few categories \cite{kim2017discoGAN,sangkloy2016scribbler}.\\\indent
We resolve the first challenge by augmenting the Sketchy database \cite{Patsorn16Sketchy}, which contains nearly 75,000 actual human \emph{sketches} paired with photos, with a larger dataset of paired \emph{edge maps} and photos. This augmentation dataset is obtained by collecting 2,299,144 Flickr images from 50 categories and synthesizing edge maps from them. During training, we adjust the ratio between edge map-image and sketch-image pairs so that the network can transfer its knowledge gradually from edge-image synthesis to sketch-image synthesis. For the second challenge, we build a GAN-based model, conditioned on an input sketch, with several additional loss terms which improve synthesis quality. We also introduce a new building block called Masked Residual Unit (MRU) which helps generate higher quality images. This block takes an extra image input and utilizes its internal mask to dynamically decide the information flow of the network. By chaining these blocks we are able to input a pyramid of images at different scales. We show that this structure outperforms naive convolutional approaches and ResNet blocks on our sketch to image synthesis tasks. 
\\
Our main contributions are:
\vspace{-.1cm}
\begin{itemize}
\item We present SketchyGAN, a deep learning approach to sketch to image synthesis. Unlike previous non-parametric approaches, we do not do image retrieval at test time. Unlike previous deep image translation methods, our network does not learn to directly copy input edges (effectively colorizing instead of converting sketches to photos). Our method is capable of generating plausible objects from 50 diverse categories. Sketch-based image synthesis is very challenging and our results are not generally \emph{photorealistic}, but we demonstrate an increase in quality compared to existing deep generative models.
\vspace{-.2cm}
\item We demonstrate a data augmentation technique for sketch data that address the lack of sufficient human-annotated training data.
\vspace{-.2cm}
\item We formulate a GAN model with additional objective functions and a new network building block. We show that all of them are beneficial for our task, and lacking any of them will reduce the quality of our results.
\end{itemize}
\begin{figure}
	\centering
	\subfloat[image][Photo]{
		\includegraphics[width=0.08\textwidth]{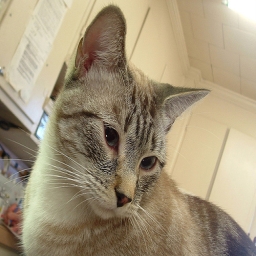}
		\label{fig:sk_em_orig}}
	\enskip
	\subfloat[Edge map][Edge~map]{
		\includegraphics[width=0.08\textwidth]{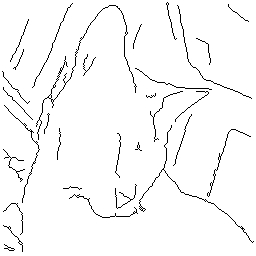}
		\label{fig:sk_em_em}}
	\enskip
	\subfloat[Sketches][Sample sketches of (a)]{
		\includegraphics[width=0.08\textwidth]{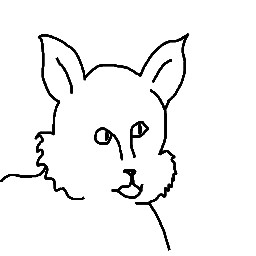}
		\includegraphics[width=0.08\textwidth]{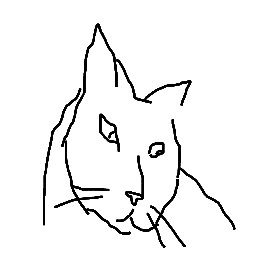}
		\includegraphics[width=0.08\textwidth]{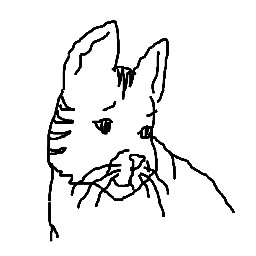}
		\label{fig:sk_em_sk}
	}
	\caption{Comparison between an edge map and sketches of the same image. The photo and sketches are from the Sketchy Database. Compared to sketches, the edge map contains more background information. The sketches, in contrast, do not precisely reflect actual object boundaries and are not spatially aligned with the object.}
	\label{fig:sk_em_comparison}
\end{figure}
\section{Related Work}
\textbf{Sketch-Based Image Retrieval and Synthesis.} 
There exist numerous works on sketch-based image retrieval \cite{eitz2010sketchFeatureDescriptor,eitz2011sketchBagOfFeatures,hu2010gradientFieldSketch,cao2011edgelSketchSearch,cao2010mindfinder,wang2010mindfinder2,hu2011bagofRegionSketch,hu2013gradientHoGSketch,james2014reenactSketchDesign,Turmukhambetov15,lin2013_3DSketchRetrieval,wang2015sketch3DRetrieval,Li2016FineGrainedSketchRetrieval}. Most methods use bag of words representations and edge detection to build features that are (ideally) invariant across both domains. Common shortcomings include the inability to perform fine-grained retrieval and the inability to map from badly drawn sketch edges to photo boundaries. To address these problems, Yu \etal \cite{Yu16SketchMe} and Sangkloy \etal \cite{Patsorn16Sketchy} train deep convolutional neural networks(CNNs) to relate sketches and photos, treating the sketch-based image retrieval as a search in the learned feature embedding space. They show that using CNNs greatly improves performance and they are able to do fine-grained and instance-level retrieval. 
Beyond the task of retrieval, Sketch2Photo \cite{chen2009sketch2photo} and PhotoSketcher \cite{Eitz11} synthesize realistic images by compositing objects and backgrounds retrieved from a given sketch. PoseShop \cite{chen2013poseshop} composites images of people by letting users input an additional 2D skeleton into the query so that the retrieval will be more precise.
\\\indent
\textbf{Sketch-Based Datasets.} 
There are only a few datasets of human-drawn sketches and they are generally small due to the effort needed to collect drawings. One of the most commonly used sketch dataset is the TU-Berlin dataset \cite{eitz2012tu_berlin} which contains 20,000 human sketches spanning 250 categories. Yu \etal \cite{Yu16SketchMe} introduced a new dataset with paired sketches and images, but there are only two categories -- shoes and chairs. There is also the CUHK Face Sketches \cite{wang2009CUHK} containing 606 face sketches drawn by artists. The newly published QuickDraw dataset \cite{ha2017SketchRNN} has an impressive 50 million sketches. However, the sketches are particularly crude because of a 10 second time limit. The sketches lack detail and tend to be iconic or canonical views. The Sketchy database~\cite{Patsorn16Sketchy}, in contrast, has more detailed drawings in a greater variety of poses. It spans 125 categories with a total of 75,471 sketches of 12,500 objects. Critically, it is the only substantial dataset of \emph{paired} sketches and photographs spanning diverse categories so we choose to use this dataset.
\\\indent
\textbf{Image-to-Image Translation with GANs.}
Generative Adversarial Networks(GANs) have shown great potential in generating natural, realistic images \cite{gulrajani2017improvedwgan,nguyen2016ppgn}. Instead of directly optimizing per pixel reconstruction error, which often leads to blurry and conservative results, GANs use a discriminator to distinguish unrealistic images from real ones thus forcing the generator to produce sharper images. 
The ``pix2pix'' work of Isola \etal \cite{pix2pix2016} demonstrates a straightforward approach to translate one image to another using conditional GANs. Conditional settings are also adapted in other image translation tasks, including sketch coloring \cite{sangkloy2016scribbler}, style transformation \cite{yoo2016pixelLevelTransfer} and domain adaptation \cite{bousmalis2016UnACGAN} tasks. In contrast with using conditional GANs and paired data, Liu \etal \cite{liu2017UNIT} introduce an unsupervised image translation framework consists of CoupledGAN \cite{liu2016CoGAN} and a pair of variational autoencoders \cite{Kingma2014VAE}. More recently, CycleGAN \cite{CycleGAN2017} shows promising results on unsupervised image translation by enforcing cycle-consistency losses.


\begin{figure}
\captionsetup[subfigure]{justification=raggedright,singlelinecheck=true,format=hang}
	\centering
	\subfloat[input][input]{
		\includegraphics[width=0.1\textwidth]{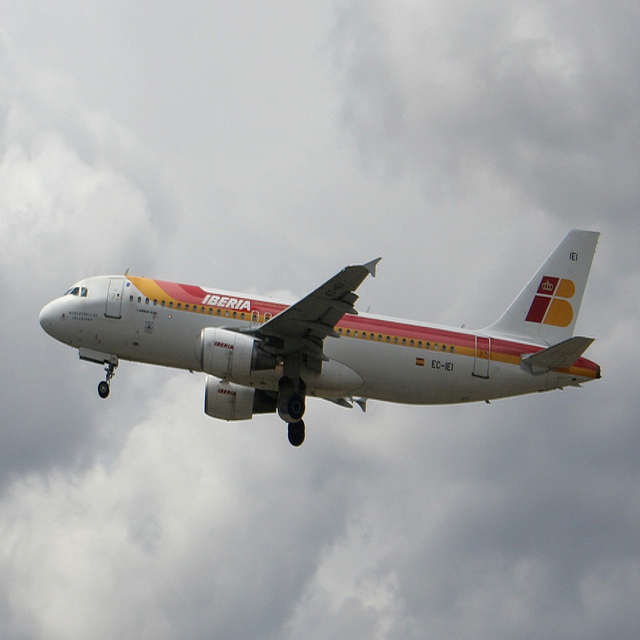}
		\label{fig:edge_map_in}}
	\enskip
	\subfloat[HED output][HED]{
		\includegraphics[width=0.1\textwidth]{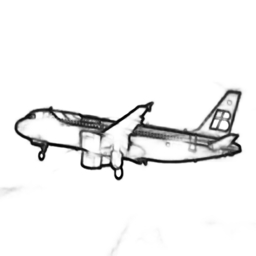}
		\label{fig:edge_map_post_0}}
	\subfloat[thinning][binarization and thinning]{
		\includegraphics[width=0.1\textwidth]{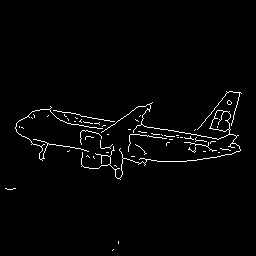}
		\label{fig:edge_map_post_2}}
    \enskip
	\subfloat[small components removing][small component removal]{
		\includegraphics[width=0.1\textwidth]{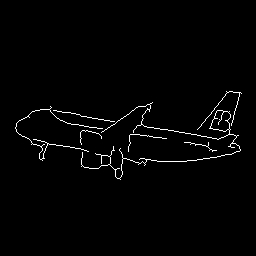}
		\label{fig:edge_map_post_3}}
	\subfloat[erosion][erosion]{
		\includegraphics[width=0.1\textwidth]{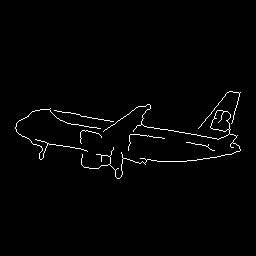}
		\label{fig:edge_map_post_4}}
	\subfloat[spurs removing][spur removal]{
		\includegraphics[width=0.1\textwidth]{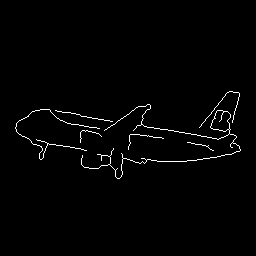}
		\label{fig:edge_map_post_5}}
	\subfloat[distance map][distance field]{
		\includegraphics[width=0.1\textwidth]{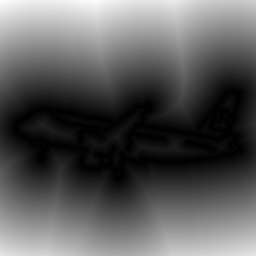}
		\label{fig:dis_map}}
	\caption{Pipeline of edge map creation. Images from intermediate steps show that each step helps remove some artifacts and make the edge maps more sketch-like.}
	\label{fig:edge_map_pipeline}
\end{figure}
\section{Sketchy Database Augmentation}
In this section, we discuss how we augment the Sketchy database~\cite{Patsorn16Sketchy} with Flickr images and synthesize edge maps which we hope approximate human sketches. The dataset is publicly available. Section \ref{Data Collection} describes image collection, image content filtering, and category selection. Section \ref{Edge Map Creation} describes our edge map synthesis. Section \ref{Training Time Scheduling} describes the way we use the augmented dataset.
\subsection{Edges vs Sketches}\label{Edge Map to Sketch}
Figure \ref{fig:sk_em_comparison} visualizes the difference between image \emph{edges} and \emph{sketches}. A sketch is set of human-drawn strokes mimicking the approximate boundary and internal contours of an object, and an edge map is machine-generated array of pixels that precisely correspond to photo intensity boundaries. Generating photos from \emph{sketches} is considerably harder than from \emph{edges}. Unlike edge maps, sketches are not precisely aligned to object boundaries, so a generative model needs to learn spatial transformations to correct deformed strokes. Second, edge maps usually contain more information about backgrounds and details, while sketches do not, so a generative model must insert more information itself. Finally, sketches may contain caricatured or iconic features, like the ``tiger'' stripes on the cat's face in Figure \ref{fig:sk_em_sk}, which a model must learn to handle. Despite these considerable differences, edge maps are still a valuable augmentation to the limited Sketchy database. 

\subsection{Data Collection}\label{Data Collection}
Learning the mapping between edges or sketches to photos requires significant training data. We want thousands of images per category. ImageNet only has around 1,000 images per class, and photos in COCO tend to be cluttered and thus not ideal as object sketch exemplars. Ideally we want photographs with one dominant object as is the case for the Sketchy database photographs. Accordingly, we collect images directly from Flickr through the Flickr API by querying category names as keywords. 100,000 images are gathered for each category, sorted by ``relevance''. Two different models are used for filtering out unrelated images. We use an Inception-ResNet-v2 network \cite{szegedy2017inceptionResNet} to filter images from the 38 ImageNet~\cite{ImageNet15} categories that overlap with Sketchy, and a Single Shot MultiBox Detector \cite{liu2016ssd} to detect whether an image contains an object in the 18 COCO~\cite{lin2014MSCOCO} categories that overlap with Sketchy. For SSD, the bounding box of a detected object must cover more than 5\% of the image area or the image is discarded. After filtering, we obtain a dataset with an average of 46,265 images per ImageNet category and 61,365 images per COCO category. For the remainder of the paper, we use 50 out of the 56 available categories after excluding six categories that often have a human as a main object. The excluded classes are harp, violin, umbrella, saxophone, racket, and trumpet.

\begin{figure}
	\centering
	\subfloat{
		\includegraphics[width=0.07\textwidth]{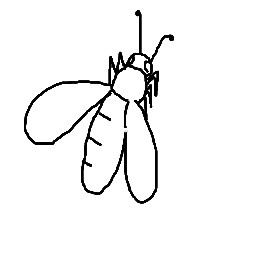}}
	\quad
	\subfloat{
		\includegraphics[width=0.35\textwidth]{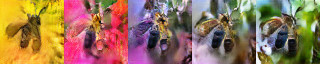}	}
	\caption{Images synthesized from the same input sketch with different noise vectors. The network learned to change a significant portion of the image (the flower), which is not conditioned by the input sketch. In each case, the bee remains plausible.}
	\label{fig:div_loss}
\end{figure}


\subsection{Edge Map Creation}\label{Edge Map Creation}
We use edge detection and several post-processing steps to obtain sketch-like edge maps. The pipeline is illustrated in Figure \ref{fig:edge_map_pipeline}. The first step is to detect edges with Holistically-nested edge detection (HED) \cite{xie2015HED} as in Isola \etal \cite{pix2pix2016}. After binarizing the output and thinning all edges \cite{zhang1984fastThinning}, we clean isolated pixels and remove small connected components. Next we perform erosion with a threshold on all edges, further decreasing number of edge fragments. Remaining spurs are then removed. Because edges are very sparse, we calculate an unsigned euclidean distance field for each edge map to obtain a dense representation (see Figure \ref{fig:dis_map}). Similar distance-field representations are used in recent works on 3D shape recovery \cite{Nguyen20163DShapeRepair,han20173DShapeCompletion}. We also calculate distance fields for sketches in the Sketchy database.
\begin{figure}
	\centering
	\includegraphics[width=0.45\textwidth]{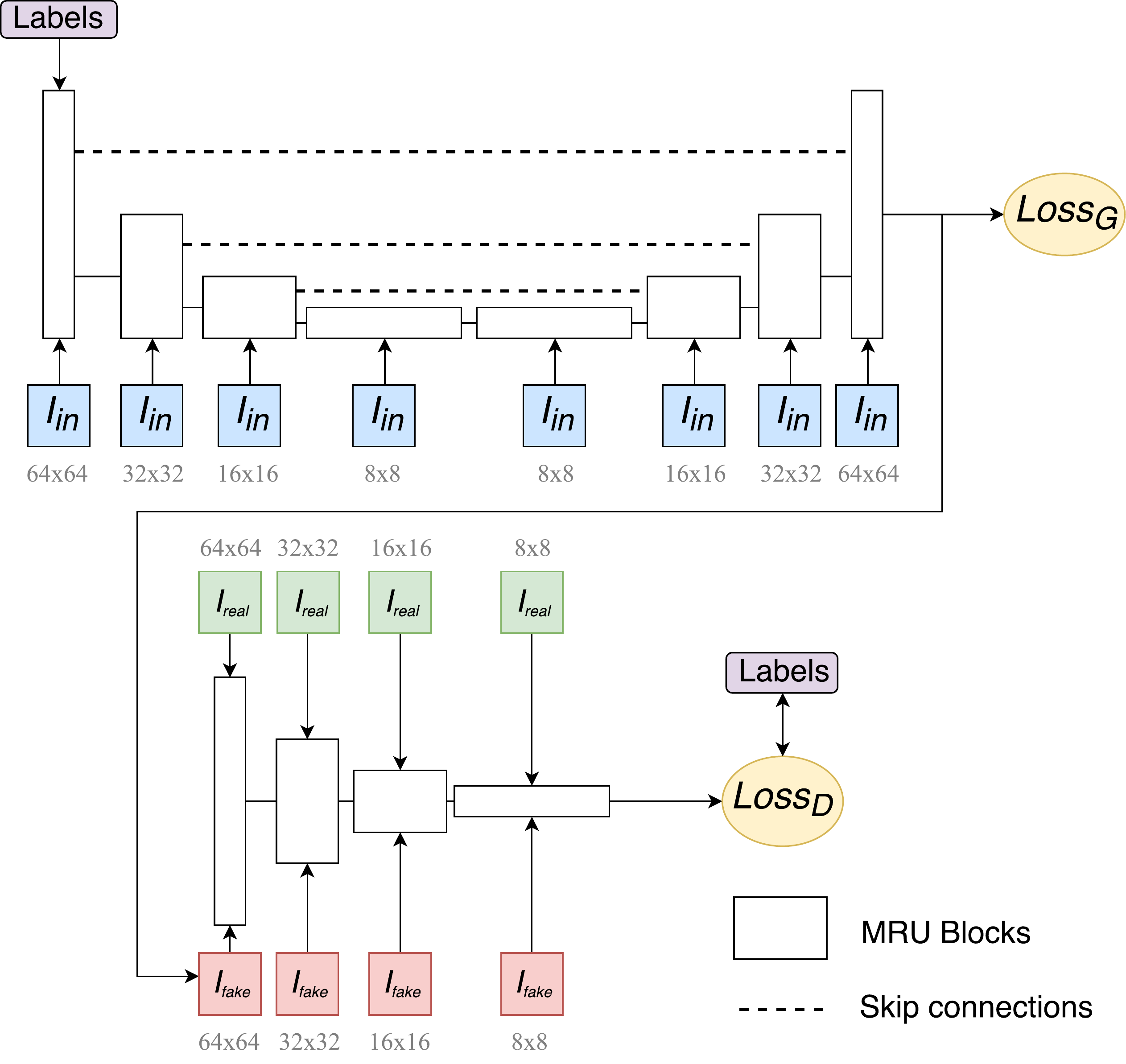}
	\caption{Complete structure of our network. Since we are using MRU blocks, both the generator and the discriminator can take multi-scale inputs.}
	\label{fig:network_structure}
\end{figure}

\subsection{Training Adaptation from Edges to Sketches}\label{Training Time Scheduling}
Because our final goal is a network that generates images from sketches, it is necessary to train the network on both edge maps and sketches. To simplify training process, we use a strategy that gradually shifts the inputs from edge maps to sketches: at the beginning of training, the training data are mostly pairs of images and edge maps. During training, we slowly increase the proportion of sketch-image pairs. Let $i_{max}$ be the maximum number of training iterations, $i_{cur}$ be the number of current iteration, then the proportion of sketches and edge maps at current iteration is given by:
\begin{equation}
P_{sk}=0.1 + min(0.8,\ ({\frac{i_{cur}}{i_{max}}})^\lambda)
\end{equation}
\begin{equation}
P_{edge}=1-P_{sk}
\end{equation}
respectively, where $\lambda$ is an adjustable hyperparameter indicating how fast the portion of sketches grows. We use $\lambda=1$ in our experiments. It is easy to see that $P_{sk}$ grows from 0.1 slowly to 0.9. Using this training schedule, we eliminate the need of separate pre-training on edge maps, so the whole training process is unified. 
We compare this method to training on edge maps first then fine-tuning on sketches. We find that discrete pre-training and then fine-tuning leads to lower inception scores on the test set compared to a gradual ramp from edges to sketches (6.73 vs 7.90).
\begin{figure}
	\centering
	\includegraphics[width=0.4\textwidth]{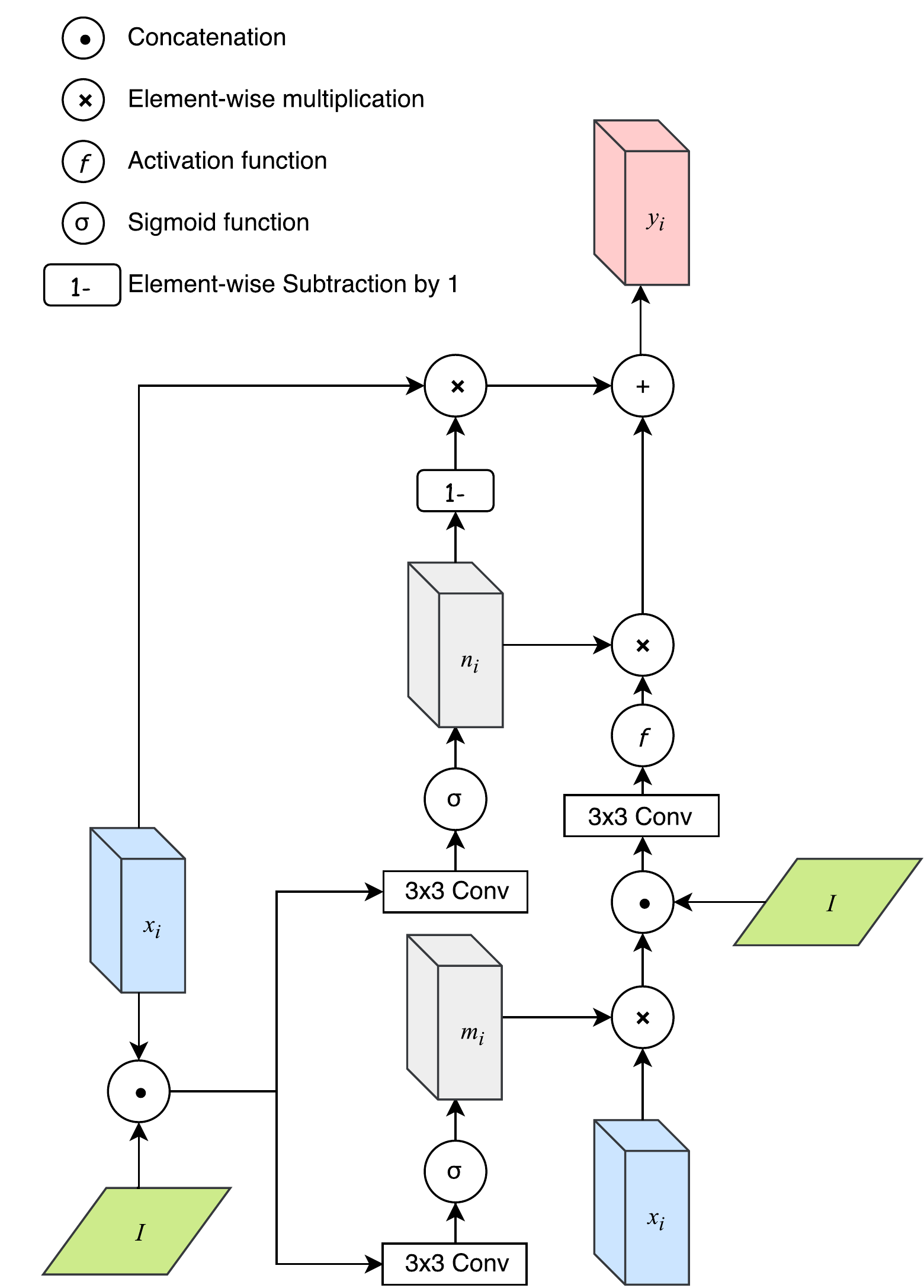}
	\caption{Structure of a Masked Residual Unit (MRU). It takes in feature maps $x_i$ and an extra image $I$, then outputs new feature maps $y_i$.}
	\label{fig:mru_structure}
\end{figure}
\section{SketchyGAN}
In this section we present a Generative Adversarial Network framework that transforms input sketches into images. Our GAN learns a mapping from an input sketch $x$ to an output image $y$, so that $G:x\rightarrow y$. The GAN has two parts, a generator $G$ and a discriminator $D$. Section \ref{MRU} introduces the Masked Residual Unit (MRU), Section \ref{Network Structure} illustrates the network structure, and Section \ref{Loss} discusses the objective functions.

\subsection{Masked Residual Unit (MRU)}\label{MRU}
We introduce a network module which allows a ConvNet to be repeatedly conditioned on an input image. The module uses a learned internal mask to selectively extract new features from the input images to combine with feature maps computed by the network thus far. We call this module the \emph{Masked Residual Unit} or \textbf{MRU}.\\\indent
\begin{figure}
	\centering
	\includegraphics[width=0.45\textwidth]{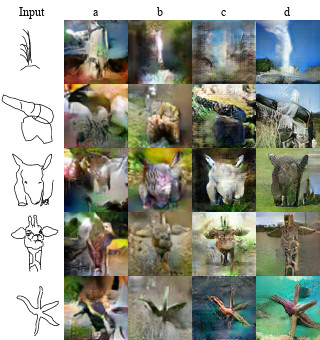}
	\caption{Image generated by pix2pix variations and our method. The four columns labeled by \textit{a} to \textit{d} are: (a) pix2pix on Sketchy (b) pix2pix on Augmented Sketchy (c) Label-supervised pix2pix on Augmented Sketchy and (d) our method. Comparing to our method, pix2pix results are blurry and noisy, often containing color patches and unwanted artifacts.}
	\label{fig:pix2pix_comparison}
\end{figure}
Figure \ref{fig:mru_structure} shows the structure of Masked Residual Unit (MRU). Qualitative and quantitative comparison to DCGAN \cite{radford15} and ResNet generative architectures can be found in Section \ref{Component Analysis}. An MRU block takes two inputs: input feature maps $x_i$ and an image $I$, and outputs feature maps $y_i$. For convenience we only discuss the case in which inputs and outputs have the same spacial dimension. Let $[\cdot,\cdot]$ denote concatenation, $Conv(x)$ denote convolution on $x$, and $f(x)$ be an activation function. We want to first merge the information in input image $I$ into input feature maps $x_i$. A naive approach will be concatenating them along the feature depth dimension and performing convolution:
\begin{equation}
{z_i}=f(Conv([x_i,\ I]))
\end{equation}
However it is better if the block can decide how much information it wants to preserve upon receiving the new image. So instead we use the following approach:
\begin{equation}
{z_i}=f(Conv([m_i\odot x_i,\ I]))
\end{equation}
where
\begin{equation}
m_i=\sigma(Conv([x_i,\ I]))
\end{equation}
is a mask over the input feature maps. Multiple convolutional layers can be stacked here to increase performance. We then want to dynamically combine the information from the newly convolved feature maps and the original input feature maps, so we use another mask
\begin{equation}
n_i=\sigma(Conv([x_i,\ I]))
\end{equation}
to combine the input feature maps with the new feature maps to get the final output:
\begin{equation}\label{y_i}
y_i=(1-n_i)\odot z_i + n_i\odot x_i
\end{equation}
The second term in Equation \ref{y_i} serves as a residual connection. Because there are internal masks to determine information flow, we call this structure masked residual unit. We can stack multiple of these units and input the same image at different scales repetitively so that the network can retrieve information from the input image dynamically on its computation path.\\\indent
The MRU formulation is similar to that of the Gated Recurrent Unit (GRU) \cite{cho2014GRU}. However, we are driven by  different motivations and there are several crucial differences: 1) We are motivated by repetitively inputting the same image to improve the information flow. GRU is designed to address vanishing gradients in recurrent neural networks. 2) GRU cells are recurrent so part of the output is fed back into the same cell, while MRU blocks are cascaded so the outputs of a previous block are fed into the next block. 3) GRU shares weights for each step so it can only receive fixed length inputs. No two MRU blocks share weights, so we can shrink or expand the size of output feature maps like normal convolutional layers.\\\indent
\begin{table}
	\centering
	\begin{tabular}{ l r }
		Model                           & Inception Score \\
		\hline
		pix2pix, Sketchy only           & 3.94 \\
		pix2pix, Augmented              & 4.53 \\
		pix2pix, Augmented+Label        & 5.49 \\
		\textbf{Ours}                   & \textbf{7.90} \\
		\hline
		Real Image                      & 15.46 \\
	\end{tabular}
    \vspace{-.1in}
	\caption{Comparison of our method to baselines methods. We compared to three variants of pix2pix, and our method shows a much higher score on test images.}
	\label{table:baseline}
\end{table}
\subsection{Network Structure}\label{Network Structure}
Our complete network structure is shown in Figure \ref{fig:network_structure}. The generator uses an encoder-decoder structure. Both the encoder and the decoder are built with MRU blocks, where the sketches are resized and fed into every MRU block on the path. In our best results in Figure \ref{fig:final_output}, we also apply skip-connections between encoder and decoder blocks, so the output feature maps from encoder blocks will be concatenated to the outputs of corresponding decoder blocks. The discriminator is also built with MRU blocks but will shrink in spatial dimension. At the end of the discriminator, we output two logits, one for the GAN loss and one for classification loss.
\subsection{Objective Function}\label{Loss}
Let $x$, $y$ be either an image or a sketch, $z$ be a noise vector, and $c$ be a class label, Our GAN objective function can be expressed as
\begin{align*}
\mathcal{L}_{GAN}(D,G)=&\mathbb{E}_{y\sim P_{image}}[\mathrm{log}\ D(y)]+\\
&\mathbb{E}_{x\sim P_{sketch},z\sim P_{z}}[\mathrm{log}(1-D(G(x,z)))] \numberthis \label{LossGAN_D}
\end{align*}
and the objective of generator $\mathcal{L}_{GAN}(G)$ will be to minimize the second term.\\\indent

\begin{figure}
	\centering
	\includegraphics[width=0.45\textwidth]{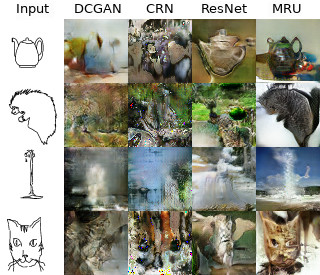}
	\caption{Visual results from DCGAN, CRN, ResNet and MRU. The MRU structure emphasize more on the main object than the other three.}
	\label{fig:mru_comparison}
\end{figure}

It is shown that giving the model side information will improve the quality of generated images \cite{odena2016ACGAN}, so we use conditional instance normalization \cite{Dumoulin17ConditionalNorm} in the generator and pass in labels of input sketches. In addition, we let the discriminator predict class labels out of the images it sees. The auxiliary classification loss of discriminator maximize the log-likelihood between predicted and ground-truth labels:
\begin{align*}
\mathcal{L}_{ac}(D)=&\mathbb{E}[\mathrm{log}\ P(C=c|y)]    \numberthis \label{LossAC_D}
\end{align*}
and the generator maximizes the same log-likelihood $\mathcal{L}_{ac}(G)=\mathcal{L}_{ac}(D)$ with discriminator fixed.\\\indent
Since we have paired image data, we are able to provide direct supervision to the network with L1-distance between generated images and ground truth images:
\begin{equation}
\mathcal{L}_{sup}(G)=\|G(x,z)-y\|_1
\end{equation}

However, directly minimizing L1 loss between generated image and ground truth image discourages diversity, so we add a perceptual loss to encourage the network to generate diverse images \cite{Dosovitskiy2016Perceptual,Johnson2016Perceptual,Chen2017CRN}. We use four intermediate layers from an Inception-V4 \cite{szegedy2017inceptionResNet} to calculate the perceptual loss. Let $\phi_i$ be the filter response of a layer in the Inception model. We define perceptual loss on the generator as:
\begin{equation}
\mathcal{L}_{p}(G)=\sum_{i}\lambda_p\|\phi_i(G(x,z))-\phi_i(y)\|_1
\end{equation}

\begin{table}
	\centering
	\begin{tabular}{ l l R{1.5cm} }
		Model                   & Num of params               & Inception Score \\
		\hline
		DCGAN                   & \textit{G}:35.1M \textit{D}:\enskip4.3M & 4.73  \\
        CRN                     & \textit{G}:21.4M \textit{D}:22.3M & 4.56  \\
		Improved ResNet         & \textit{G}:33.0M \textit{D}:31.2M & 5.76  \\
        MRU (GAN loss only)     & \textit{G}:28.1M \textit{D}:29.9M & 8.31  \\
		MRU                     & \textit{G}:28.1M \textit{D}:29.9M & 7.90  \\
		\hline
	\end{tabular}
    \vspace{-.1in}
	\caption{Comparison of MRU, CRN, ResNet and DCGAN under the same setting. DCGAN structure is included for completeness. Under similar number of parameters, MRU outperforms ResNet block significantly on our generative task.}
	\label{table:mru_resnet}
\end{table}

To further encourage diversity, we concatenate Gaussian noise to feature maps at the bottleneck of the generator. Previous works reach the conclusion that conditional GANs tend to ignore the noise completely \cite{pix2pix2016} or produce worse results because of noise \cite{pathak2016contextEncoder}. A simple diversity loss
\begin{equation}
\mathcal{L}_{div}(G)=-\lambda_{div}\|G(x,z_1)-G(x,z_2)\|_1
\end{equation}
will improve both quality and diversity of generated images. The interpretation is straightforward: with a pair of different noise vectors $z_1$ and $z_2$ conditioned on the same image, the generator should output a pair of sightly different images.\\\indent
Our complete discriminator and generator losses are thus
\begin{align*}
&\mathcal{L}(D)=\mathcal{L}_{GAN}(D,G)+\mathcal{L}_{ac}(D)    \numberthis\label{LossD}\\
&\mathcal{L}(G)=\mathcal{L}_{GAN}(G)-\mathcal{L}_{ac}(G)\\
&\qquad\quad+\mathcal{L}_{sup}(G)+\mathcal{L}_{p}(G)+\mathcal{L}_{div}(G)    \numberthis\label{LossG}\\
\end{align*}
where the discriminator maximizes Equation \ref{LossD} and the generator minimizes Equation \ref{LossG}. In practice, we use DRAGAN loss \cite{kodali2017dragan} in order to stabilize training and use focal loss \cite{Lin2017Focal} as classification loss.

\section{Experiments}\label{Experiments}
\subsection{Experiment settings}
\textbf{Dataset splitting}
We use the sketch-image pairs in selected 50 categories from training split of Sketchy as basic training data, and augment them with edge map-image pairs.
In the following sections, we call data from Sketchy Database ``Sketchy'', and Sketchy augmented with edge maps ``Augmented Sketchy''. Since we are only interested in sketch to image synthesis, all models are tested on the test split of Sketchy. All images are resized to 64$\times$64 regardless of the original aspect ratio. Both sketches and edge maps are converted into distance fields.\\\indent
\textbf{Implementation Details}
In all experiments, we use batch size of 8, except for Figure \ref{fig:final_output} which uses a batch size of 32. We use random horizontal flipping during training. We use the Adam optimizer \cite{kingma2014adam}, and set the initial learning rate of generator at 0.0001 and that of discriminator at 0.0002~\cite{heusel2017TTUR}.\\\indent
\textbf{Evaluation Metrics}
For our task of image synthesis, we use Inception Scores~\cite{salimans2016improvedGAN} to measure the quality of synthesized images. The intuition behind Inception Score is that a good synthesized image should have easily recognizable objects by an off-the-shelf recognition system. Beyond Inception Scores, we also perform a perceptual study evaluating how realistic the generated images are and how faithful they are to the input sketches.

\begin{table}
	\centering
	\begin{tabular}{ l r }
		Model                    & Input correctly identified? \\
		\hline
        Sketchy 1-NN retrieval          & 35.3\% \\
        pix2pix, Augmented+Label        & 65.9\% \\
        Ours                            & 47.4\% \\
		\hline
	\end{tabular}
    \vspace{-.1in}
	\caption{Faithfulness test on three models. Models for which participants could pick the input sketch are considered more ``faithful''.}\vspace{-0.2cm}
	\label{table:faithfulness}
\end{table}
\begin{table}
	\centering
	\begin{tabular}{ l r }
		Model                           & Picked as more realistic? \\
		\hline
        pix2pix, Sketchy only           & 6.03\% \\
		pix2pix, Augmented              & 18.4\% \\
		pix2pix, Augmented+Label        & 21.8\% \\
        Ours                            & 53.7\% \\
		\hline
	\end{tabular}
    \vspace{-.1in}
	\caption{Realism test on four generative models. We report how often results from each model were chosen by participants to be more ``realistic'' than a competing model.}
	\label{table:realism}
\end{table}
\subsection{Comparison to Baselines}
Our comparisons focus on the popular pix2pix and its variations. All models are trained for 300k iterations except for the first model.
We include three baselines:\\
\textbf{pix2pix on Sketchy} This is the simplest model. We directly take the authors' pix2pix code and train it on the 50 categories from Sketchy. Since we find the image quality stops improving after 100k iterations, we stop early at 150k iteration and report the results.\\
\textbf{pix2pix on Augmented Sketchy} In this model, we train pix2pix on both the image-edge map and image-sketch pairs, as we do in our method. The network structure and loss functions remain unchanged.\\
\textbf{Label-Supervised pix2pix on Augmented Sketchy} In this model, we modify pix2pix to pass class labels into the generator using conditional instance normalization, and also add auxiliary classification loss to its discriminator. This is a much stronger baseline, since the label information helps the network decide the object type and in turn improves the generated image quality~\cite{gulrajani2017improvedwgan,odena2016ACGAN}.\\\indent
The comparison of Inception Scores can be found in Table \ref{table:baseline} and visual results can be found in Figure \ref{fig:pix2pix_comparison}. Our observations are as follows: (i) pix2pix trained on Sketchy fails, generating unidentifiable color patches. The model is unable to translate from sketches to images. Since pix2pix has been successful with edge-to-image translations, this implies that sketch-to-image synthesis is more difficult. (ii) pix2pix trained on Augmented Sketchy performs slightly better, starting to produce the general shape of the object. This shows that edge maps help the training. (iii) The label-supervised pix2pix on Augmented Sketchy is better than the previous two baselines. It correctly colors the object more often and starts to generate some meaningful backgrounds. The results are still blurry, and many artifacts can be observed. (iv) Comparing to baselines, our method generates sharper images, gets the object color correct, puts more detailed textures on the object, and outputs meaningful backgrounds. The whole images are also more colorful.
\begin{table}
	\centering
	\begin{tabular}{ R{1.1cm} C{0.9cm} C{0.9cm} C{1.0cm} C{0.8cm} C{0.9cm} }
		\multicolumn{6}{l}{\includegraphics[width=0.4\textwidth]{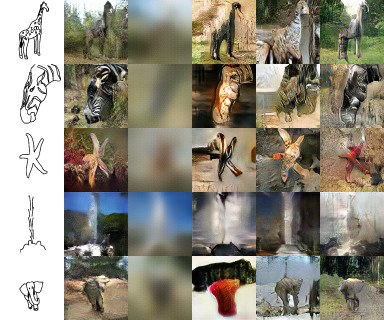}}\\ \hline
		Input      & Full  & -GAN   & -L-AC  & -P     & -DIV\\ \hline
		None       & 7.90  & 1.49   & 6.64   & 6.70   & 7.29\\
		\hline
	\end{tabular}
	\caption{Table of Inception scores for models with particular components removed. ``Full'' is the full model described in this work. ``-GAN'' means no GAN loss and no discriminator. ``-L-AC'' means no labels-supervision on generator and no auxiliary loss on discriminator. ``-P'' means no L1 and no perceptual loss, and ``-DIV'' means no diversity loss.}
	\label{table:loss_comparison}
\end{table}
\subsection{Component Analysis}\label{Component Analysis}
Here we analyze which part of our model is more important. We decouple our objective function and analyze the influence of each part of it. All models are trained on Augmented Sketchy with the same set of parameters. Detailed comparison can be found in Table \ref{table:loss_comparison}. We first remove the GAN loss and the discriminator. The result is surprisingly poor as the images are extremely vague. This observation is consistent with that of Isola \etal \cite{pix2pix2016}. Next we remove the auxiliary loss and substitute conditional instance normalization with batch normalization \cite{ioffe2015batchnorm}. This leads to a significant decrease in image quality as well as wrong colors and misplaced textures. This indicates that class information helps a lot, which makes sense because we are generating 50 categories from a single model. We then remove the L1 loss and the perceptual loss. We find they also have a large impact on image quality. From sample images we can see the model uses incorrect colors and fails and object boundaries are unrealistic or missing. Finally, we remove the diversity loss, and doing so also decreases image quality slightly. This can be related to how we apply this diversity loss, which forces the generator to generate image pairs that are realistic but different. This encourages generalization because the generator needs to find a solution that when given different noise vectors only makes changes in unconstrained areas (e.g. the background). 

\textbf{Comparison between MRU and other structures}
To demonstrate the effectiveness of our MRU blocks, we compare the performance of MRU, ResNet, Cascaded Refinement Network (CRN) \cite{Chen2017CRN} and DCGAN structures in our image synthesis task. We train several additional models: one uses improved ResNet blocks \cite{he2016improvedResNet}, which is the best variant published \cite{he2016ResNet}, in both generator and discriminator; one is a weak baseline, using DCGAN structure; one uses CRN in generator instead of MRU; and one MRU model using only GAN loss and ACGAN loss. We keep the number of parameters of MRU model and that of ResNet model roughly the same by reducing feature depth in MRU. Detailed parameter counts can be found in Table \ref{table:mru_resnet}. Judging from both visual quality and the Inception Scores, the MRU model generates better images than both ResNet and CRN models, and we show that even using only standard GAN losses, MRU outperforms other structures significantly. From Figure \ref{fig:mru_comparison}, we notice that the MRU model tends to produce higher quality foreground objects. This can be due to the internal masks of MRU serving as an attention mechanism, causing the network to selectively focus on the main object. In our task this is helpful, since we are mainly interested in generating a specific object from sketch.

\begin{figure}[t]
	\centering
	\subfloat{
		\includegraphics[width=0.13\textwidth]{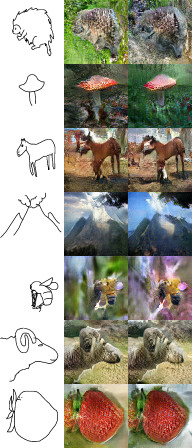}
		\label{fig:final_output_1}}
	\subfloat{
		\includegraphics[width=0.13\textwidth]{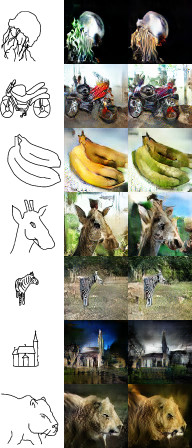}
		\label{fig:final_output_2}}
	\subfloat{
		\includegraphics[width=0.13\textwidth]{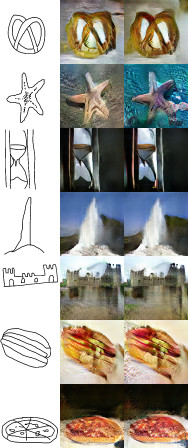}
		\label{fig:final_output_3}}
	\caption{Some of the best output images from our full model. For each input sketch, we show a pair of output images to demonstrate the diversity of our model.}
	\label{fig:final_output}
\end{figure}
\subsection{Human Evaluation of Realism and Faithfulness}\label{Human Evaluation}
We do two human evaluations to measure how our model compares against baselines in terms of realism and faithfulness to the input sketch. In the ``faithfulness'' test, a participant sees the output of either pix2pix, SketchyGAN or 1-nearest-neighbor retrieval using the representation learned in the Sketchy Database \cite{Patsorn16Sketchy}. With each image, the participant also sees 9 random sketches of the same category, one of which is the actual input/query sketch. The participant is asked to pick the sketch that prompted the output image. We then count how often participants pick the correct input sketch, so a higher correct selection rate indicates the model produces a more ``faithful'' output. In the ``realism'' test, a participant sees the output of pix2pix variants and SketchyGAN compared in pairs, alongside the corresponding input sketch. The participant is asked to pick the image that they think is more realistic. For each model we calculate how often participants think it is more realistic. The image retrieval baseline is not evaluated for realism since it only returns existing, realistic photographs. We conducted 696 trails for the ``faithfulness'' test and 348 trails for the ``realism'' test. The results show that SketchyGAN is more faithful than the retrieval model, but is less faithful than pix2pix which often preserves the input edges precisely (Table~\ref{table:faithfulness}). Meanwhile, SketchyGAN is considered more realistic than pix2pix variants (Table~\ref{table:realism}). The results are consistent with our goal that our model should respect the intent of input sketches, but at the same time deviate from the strokes if necessary in order to produce realistic images.

\section{Conclusion}\label{Conclusion}
In this work, we presented a novel approach to the sketch-to-image synthesis problem. The problem is challenging given the nature of sketches, and this introduced a deep generative model that is promising in sketch to image synthesis. We introduced a data augmentation technique for sketch-image pairs to encourage research in this direction. The demonstrated GAN framework can synthesize more realistic images than popular generative models, and the generated images are diverse. Currently, the main focus on GANs is to find better probability metrics as objective functions, but there has been very few works searching for better network structures in GANs. We proposed a new network structure for our generative task, and we showed that it performs better than existing structures.

\textbf{Limitations.} Ideally, we want our results to be both realistic \emph{and} faithful to the \textbf{intent} of the input sketch. For many sketches, we fail to meet one or both of these goals. Results generally aren't photorealistic, nor are they high enough resolution. Sometimes realism is lost by being \emph{overly} faithful to the sketch -- e.g. Skinny horse legs that too closely follow the badly drawn input boundaries (Figure~\ref{fig:final_output}). In other cases, we do deviate from the user sketch to make the output more realistic (motorcycle and plane in Figure~\ref{fig:final_output_abs}, mushroom, church, geyser, and castle in Figure~\ref{fig:final_output}) but still respect the pose and position of the object in the input sketch. This is more desirable. Human \textbf{intent} is hard to learn, and SketchyGAN failures that treat the input sketch too literally may be due to lack of sketch-photo training pairs. Despite the fact that our results are not yet \textit{photorealistic}, we think they show a substantial improvement over previous methods.

\textbf{Acknowledgements.} This work was funded by NSF award 1561968.

\newpage

{\small
\bibliographystyle{ieee}
\bibliography{main}
}

\clearpage

\section*{Supplementary Material Outline}
\setcounter{section}{0}
\renewcommand*{\theHsection}{chSupp.\the\value{section}}

\noindent Section \ref{cats} lists all categories we used in training our models. Section \ref{clsf} compares the performance of MRU to some other models on CIFAR-10. Section \ref{figs} shows samples of generated images from all 50 categories.

\section{Category list}\label{cats}

Here are the 50 categories we use for training and testing our models:
airplane,
ant,
apple,
banana,
bear,
bee,
bell,
bench,
bicycle,
candle,
cannon,
car,
castle,
cat,
chair,
church,
couch,
cow,
cup,
dog,
elephant,
geyser,
giraffe,
hammer,
hedgehog,
horse,
hotdog,
hourglass,
jellyfish,
knife,
lion,
motorcycle,
mushroom,
pig,
pineapple,
pizza,
pretzel,
rifle,
scissors,
scorpion,
sheep,
snail,
spoon,
starfish,
strawberry,
tank,
teapot,
tiger,
volcano,
zebra.

\section{Evaluation of MRU on CIFAR-10}\label{clsf}
We introduce the Masked Residual Unit (MRU) to improve generative deep networks by giving repeated access to the conditioning signal (in our case, a sketch). But this network building block is also quite useful for classification tasks. We compare the performance of the MRU and other recent architectures on CIFAR-10 and show that the MRU performance is on par with ResNet. Accuracy numbers for other models are obtained from their corresponding papers. For convenience, we call the improved ResNet "ResNet-v2" in the table. In "MRU-108, LeakyReLU gate", we substitute the sigmoid activations in our MRU units with LeakyReLU \cite{Maas13LReLU}, and normalize obtained masks to the range of $[0, 1]$.
\begin{table}[h]
	\centering
	\begin{tabular}{ l r }
		Model                                         & error (\%) \\
		\hline
		NIN \cite{lin2013NIN}                         & 8.81 \\
		Highway \cite{srivastava2015Highway}          & 7.72 \\
		ResNet-110 \cite{he2016ResNet}                & 6.61 \\
		ResNet-1202 \cite{he2016ResNet}               & 7.93 \\
		ResNet-v2-164 \cite{he2016improvedResNet}     & 5.46 \\
		\hline
		MRU-108                              & 6.34 \\
		\textbf{MRU-108, LeakyReLU gate}                  & \textbf{5.83} \\
		\hline
	\end{tabular}
	\caption{Comparison of error rates on CIFAR-10. Lower is better.}
	\label{table:baseline_supp}
\end{table}

\section{Samples from all 50 categories}\label{figs}
Here we present samples from all 50 categories from pix2pix variants and our methods for comparison. Each category contains three input samples, among which the third sample is a failure case for our method. The six columns in each figure are: (Input) input sketch, (a) pix2pix on Sketchy, (b) pix2pix on Augmented Sketchy, (c) Label-supervised pix2pix on Augmented Sketchy, (d) our method, (GT) ground truth image.

\begin{figure}[!htb]
	\centering
	\quad\textbf{airplane}\par\medskip
	\includegraphics[width=0.5\textwidth]{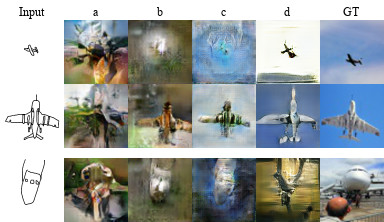}
\end{figure}

\begin{figure}[!htb]
	\centering
	\quad\textbf{ant}\par\medskip
	\includegraphics[width=0.5\textwidth]{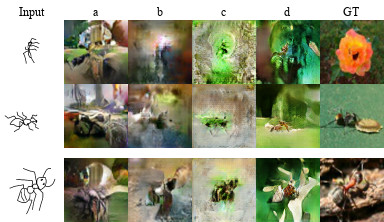}
\end{figure}

\begin{figure}[!htb]
	\centering
	\quad\textbf{apple}\par\medskip
	\includegraphics[width=0.5\textwidth]{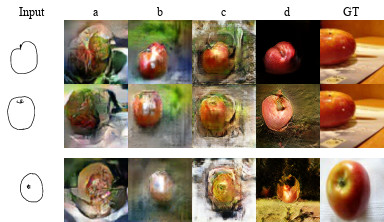}
\end{figure}

\begin{figure}[!htb]
	\centering
	\quad\textbf{banana}\par\medskip
	\includegraphics[width=0.5\textwidth]{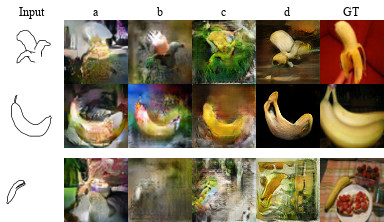}
\end{figure}

\begin{figure}[!htb]
	\centering
	\quad\textbf{bear}\par\medskip
	\includegraphics[width=0.5\textwidth]{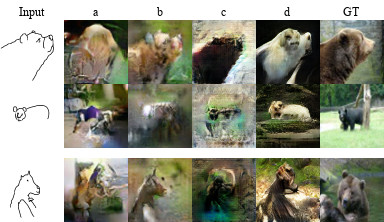}
\end{figure}

\begin{figure}[!htb]
	\centering
	\quad\textbf{bee}\par\medskip
	\includegraphics[width=0.5\textwidth]{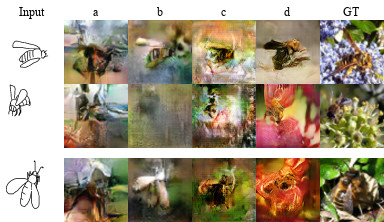}
\end{figure}

\begin{figure}[!htb]
	\centering
	\quad\textbf{bell}\par\medskip
	\includegraphics[width=0.5\textwidth]{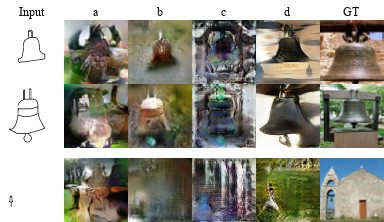}
\end{figure}

\begin{figure}[!htb]
	\centering
	\quad\textbf{bench}\par\medskip
	\includegraphics[width=0.5\textwidth]{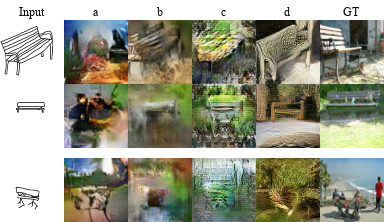}
\end{figure}

\begin{figure}[!htb]
	\centering
	\quad\textbf{bicycle}\par\medskip
	\includegraphics[width=0.5\textwidth]{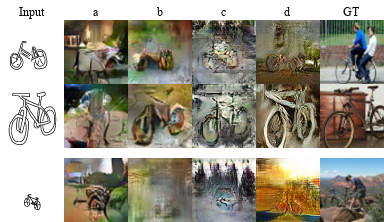}
\end{figure}

\begin{figure}[!htb]
	\centering
	\quad\textbf{candle}\par\medskip
	\includegraphics[width=0.5\textwidth]{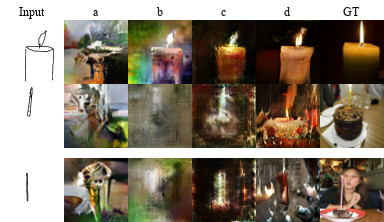}
\end{figure}

\begin{figure}[!htb]
	\centering
	\quad\textbf{cannon}\par\medskip
	\includegraphics[width=0.5\textwidth]{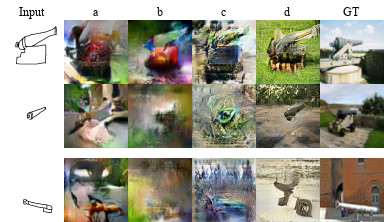}
\end{figure}

\begin{figure}[!htb]
	\centering
	\quad\textbf{car}\par\medskip
	\includegraphics[width=0.5\textwidth]{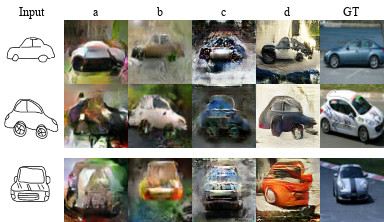}
\end{figure}

\begin{figure}[!htb]
	\centering
	\quad\textbf{castle}\par\medskip
	\includegraphics[width=0.5\textwidth]{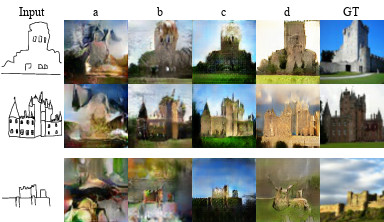}
\end{figure}

\begin{figure}[!htb]
	\centering
	\quad\textbf{cat}\par\medskip
	\includegraphics[width=0.5\textwidth]{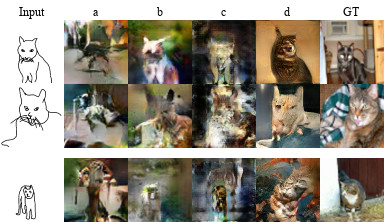}
\end{figure}

\begin{figure}[!htb]
	\centering
	\quad\textbf{chair}\par\medskip
	\includegraphics[width=0.5\textwidth]{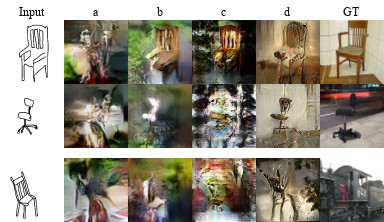}
\end{figure}

\begin{figure}[!htb]
	\centering
	\quad\textbf{church}\par\medskip
	\includegraphics[width=0.5\textwidth]{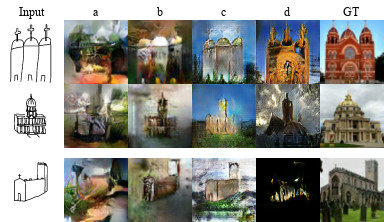}
\end{figure}

\begin{figure}[!htb]
	\centering
	\quad\textbf{couch}\par\medskip
	\includegraphics[width=0.5\textwidth]{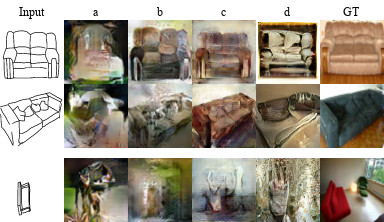}
\end{figure}

\begin{figure}[!htb]
	\centering
	\quad\textbf{cow}\par\medskip
	\includegraphics[width=0.5\textwidth]{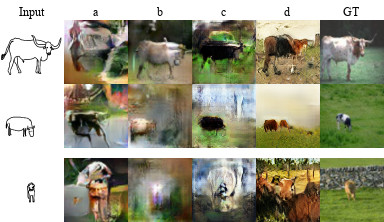}
\end{figure}

\begin{figure}[!htb]
	\centering
	\quad\textbf{cup}\par\medskip
	\includegraphics[width=0.5\textwidth]{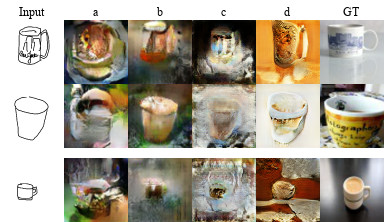}
\end{figure}

\begin{figure}[!htb]
	\centering
	\quad\textbf{dog}\par\medskip
	\includegraphics[width=0.5\textwidth]{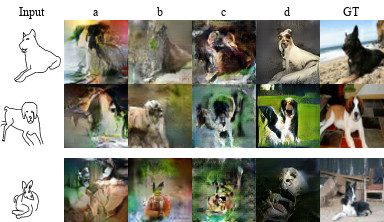}
\end{figure}

\begin{figure}[!htb]
	\centering
	\quad\textbf{elephant}\par\medskip
	\includegraphics[width=0.5\textwidth]{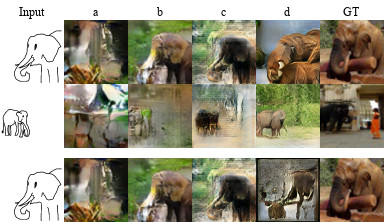}
\end{figure}

\begin{figure}[!htb]
	\centering
	\quad\textbf{geyser}\par\medskip
	\includegraphics[width=0.5\textwidth]{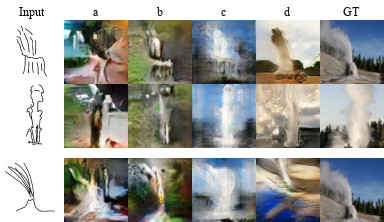}
\end{figure}

\begin{figure}[!htb]
	\centering
	\quad\textbf{giraffe}\par\medskip
	\includegraphics[width=0.5\textwidth]{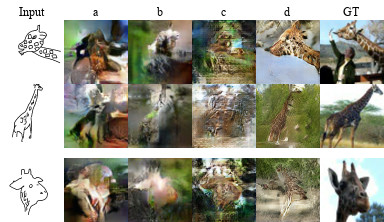}
\end{figure}

\begin{figure}[!htb]
	\centering
	\quad\textbf{hammer}\par\medskip
	\includegraphics[width=0.5\textwidth]{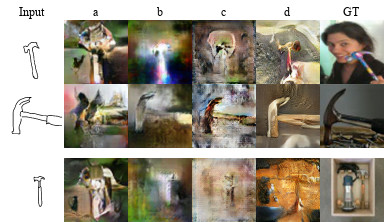}
\end{figure}

\begin{figure}[!htb]
	\centering
	\quad\textbf{hedgehog}\par\medskip
	\includegraphics[width=0.5\textwidth]{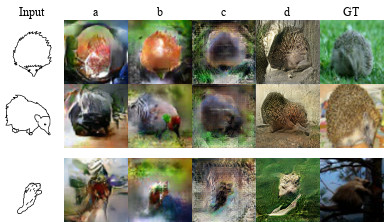}
\end{figure}

\begin{figure}[!htb]
	\centering
	\quad\textbf{horse}\par\medskip
	\includegraphics[width=0.5\textwidth]{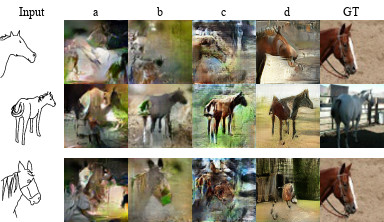}
\end{figure}

\begin{figure}[!htb]
	\centering
	\quad\textbf{hotdog}\par\medskip
	\includegraphics[width=0.5\textwidth]{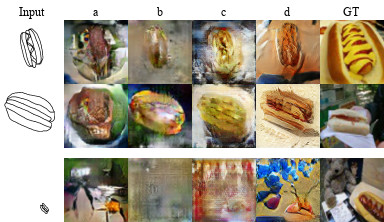}
\end{figure}

\begin{figure}[!htb]
	\centering
	\quad\textbf{hourglass}\par\medskip
	\includegraphics[width=0.5\textwidth]{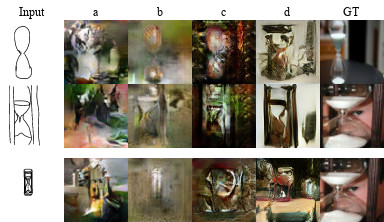}
\end{figure}

\begin{figure}[!htb]
	\centering
	\quad\textbf{jellyfish}\par\medskip
	\includegraphics[width=0.5\textwidth]{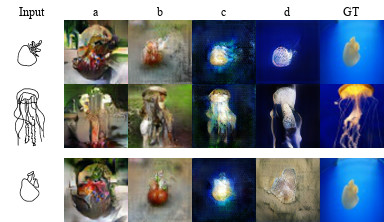}
\end{figure}

\begin{figure}[!htb]
	\centering
	\quad\textbf{knife}\par\medskip
	\includegraphics[width=0.5\textwidth]{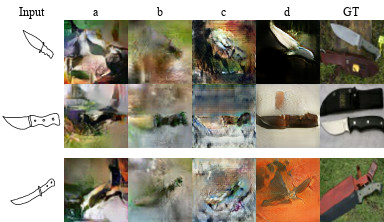}
\end{figure}

\begin{figure}[!htb]
	\centering
	\quad\textbf{lion}\par\medskip
	\includegraphics[width=0.5\textwidth]{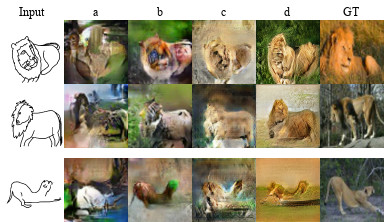}
\end{figure}

\begin{figure}[!htb]
	\centering
	\quad\textbf{motorcycle}\par\medskip
	\includegraphics[width=0.5\textwidth]{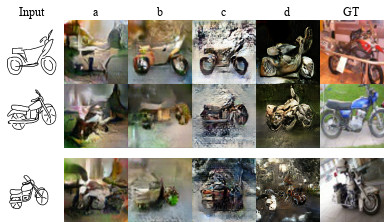}
\end{figure}

\begin{figure}[!htb]
	\centering
	\quad\textbf{mushroom}\par\medskip
	\includegraphics[width=0.5\textwidth]{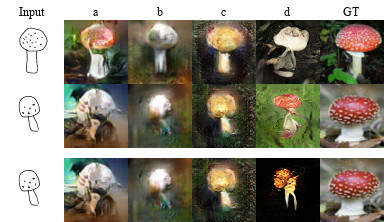}
\end{figure}

\begin{figure}[!htb]
	\centering
	\quad\textbf{pig}\par\medskip
	\includegraphics[width=0.5\textwidth]{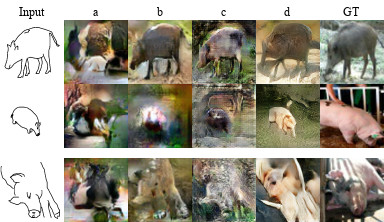}
\end{figure}

\begin{figure}[!htb]
	\centering
	\quad\textbf{pineapple}\par\medskip
	\includegraphics[width=0.5\textwidth]{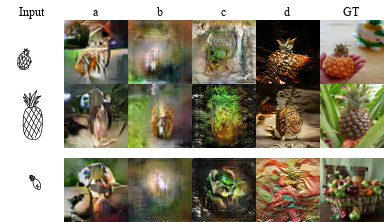}
\end{figure}

\begin{figure}[!htb]
	\centering
	\quad\textbf{pizza}\par\medskip
	\includegraphics[width=0.5\textwidth]{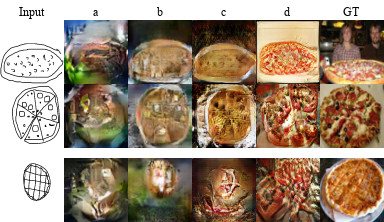}
\end{figure}

\begin{figure}[!htb]
	\centering
	\quad\textbf{pretzel}\par\medskip
	\includegraphics[width=0.5\textwidth]{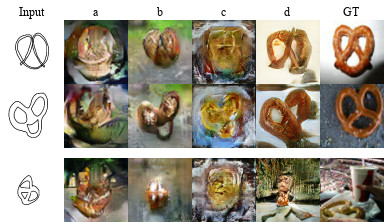}
\end{figure}

\begin{figure}[!htb]
	\centering
	\quad\textbf{rifle}\par\medskip
	\includegraphics[width=0.5\textwidth]{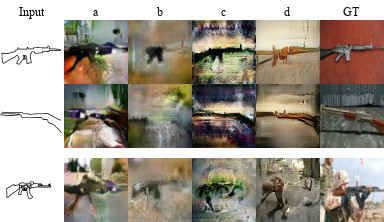}
\end{figure}

\begin{figure}[!htb]
	\centering
	\quad\textbf{scissors}\par\medskip
	\includegraphics[width=0.5\textwidth]{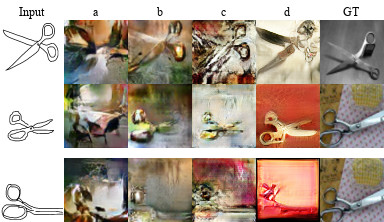}
\end{figure}

\begin{figure}[!htb]
	\centering
	\quad\textbf{scorpion}\par\medskip
	\includegraphics[width=0.5\textwidth]{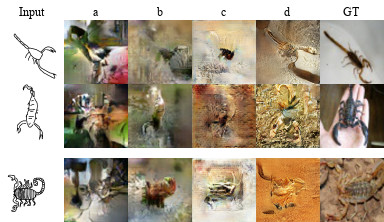}
\end{figure}

\begin{figure}[!htb]
	\centering
	\quad\textbf{sheep}\par\medskip
	\includegraphics[width=0.5\textwidth]{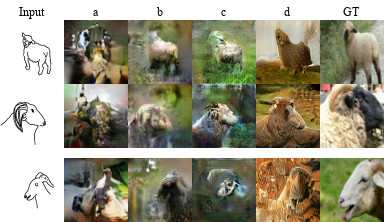}
\end{figure}

\begin{figure}[!htb]
	\centering
	\quad\textbf{pig}\par\medskip
	\includegraphics[width=0.5\textwidth]{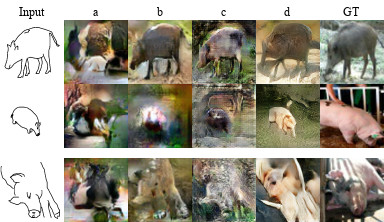}
\end{figure}

\begin{figure}[!htb]
	\centering
	\quad\textbf{snail}\par\medskip
	\includegraphics[width=0.5\textwidth]{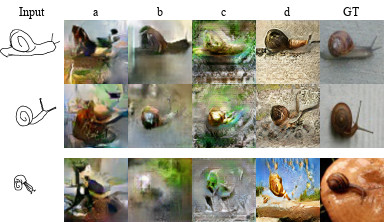}
\end{figure}

\begin{figure}[!htb]
	\centering
	\quad\textbf{spoon}\par\medskip
	\includegraphics[width=0.5\textwidth]{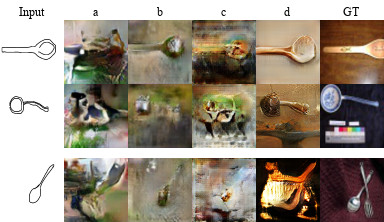}
\end{figure}

\begin{figure}[!htb]
	\centering
	\quad\textbf{starfish}\par\medskip
	\includegraphics[width=0.5\textwidth]{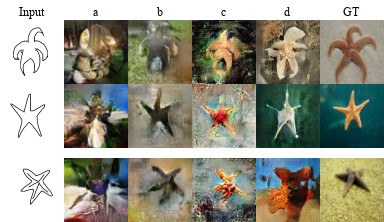}
\end{figure}

\begin{figure}[!htb]
	\centering
	\quad\textbf{strawberry}\par\medskip
	\includegraphics[width=0.5\textwidth]{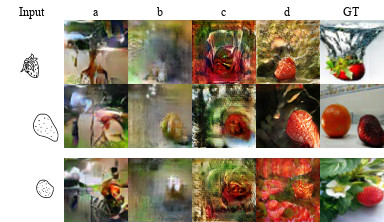}
\end{figure}

\begin{figure}[!htb]
	\centering
	\quad\textbf{tank}\par\medskip
	\includegraphics[width=0.5\textwidth]{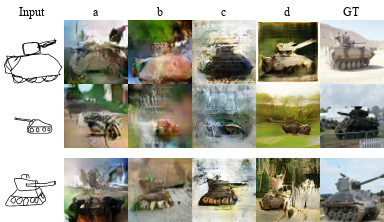}
\end{figure}

\begin{figure}[!htb]
	\centering
	\quad\textbf{teapot}\par\medskip
	\includegraphics[width=0.5\textwidth]{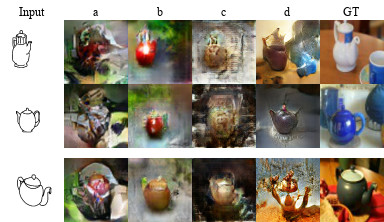}
\end{figure}

\begin{figure}[!htb]
	\centering
	\quad\textbf{tiger}\par\medskip
	\includegraphics[width=0.5\textwidth]{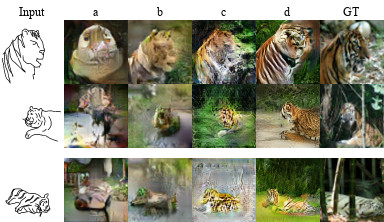}
\end{figure}

\begin{figure}[!htb]
	\centering
	\quad\textbf{volcano}\par\medskip
	\includegraphics[width=0.5\textwidth]{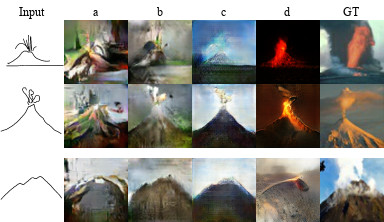}
\end{figure}

\begin{figure}[!htb]
	\centering
	\quad\textbf{zebra}\par\medskip
	\includegraphics[width=0.5\textwidth]{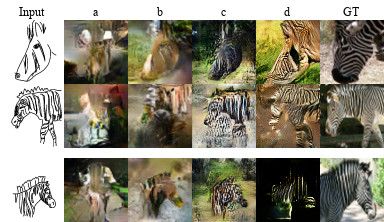}
\end{figure}

\clearpage

\end{document}